\documentclass[letterpaper, 10 pt, conference]{ieeeconf}  

\IEEEoverridecommandlockouts                              

\usepackage{graphicx} 
\usepackage{amsmath} 
\usepackage{cite}
\usepackage{color,soul}
\usepackage[caption=false]{subfig}
\usepackage{lipsum}
\usepackage{hyperref}
\title{\LARGE \bf
Step Timing Adjustment: A Step toward Generating Robust Gaits *
}

\author{Majid Khadiv$^{1,2}$, Alexander Herzog$^{2}$, S. Ali. A. Moosavian$^{1}$, and Ludovic Righetti$^{2}$
\thanks{*This research was supported by the Max-Planck Society, MPI-ETH center for learning systems and the European Research Council under the European Union's Horizon 2020 research and innovation program (grant agreement No 637935).
}
\thanks{$^{1}$Department of Mechanical Engineering, K. N. Toosi University of Technology, Tehran, Iran{\tt\small (mkhadiv@mail.kntu.ac.ir)}
{\tt\small (moosavian@kntu.ac.ir)}.}
\thanks{$^{2}$Autonomous Motion Department, Max-Planck Institute for Intelligent Systems, Germany.
{\tt\small (majid.khadiv@tuebingen.mpg.de)}
{\tt\small (alexander.herzog@tuebingen.mpg.de)}
{\tt\small (ludovic.righetti@tuebingen.mpg.de)}.}      
}

\begin{document}

\maketitle
\thispagestyle{empty}
\pagestyle{empty}

\begin{abstract}

Step adjustment for humanoid robots has been shown to improve robustness in gaits. However, step duration adaptation is often neglected in control strategies. In this paper, we propose an approach that combines both step location and timing adjustment for generating robust gaits. In this approach, step location and step timing are decided, based on feedback from the current state of the robot. The proposed approach is comprised of two stages. In the first stage, the nominal step location and step duration for the next step or a previewed number of steps are specified. In this stage which is done at the start of each step, the main goal is to specify the best step length and step duration for a desired walking speed. The second stage deals with finding the best landing point and landing time of the swing foot at each control cycle. In this stage, stability of the gaits is preserved by specifying a desired offset between the swing foot landing point and the Divergent Component of Motion (DCM) at the end of current step. After specifying the landing point of the swing foot at a desired time, the swing foot trajectory is regenerated at each control cycle to realize desired landing properties. Simulation on different scenarios shows the robustness of the generated gaits from our proposed approach compared to the case where no timing adjustment is employed.

\end{abstract}

\section{INTRODUCTION}

Generating walking patterns exploiting the whole dynamics of humanoid robots is a general approach which
can deal with multi-contact interaction with environment and walking on different surfaces \cite{herzog2015trajectory,carpentier2016A,khadiv2015optimal}. However solving a high dimensional nonlinear optimization demands high computation burden. Furthermore, due to the non-convexity of the problem, convergence to the global minimum is not guaranteed. As a result, simplified linear models that capture the task relevant dynamics to a set of linear equations are useful for generating walking patterns in real-time. In this context, the Linear Inverted Pendulum Model (LIPM) \cite{kajita20013d}, has been very successfully used for the design of walking controllers for complex biped robots. Kajita et al. \cite{kajita2003biped} proposed preview control of the Zero Moment Point (ZMP) to generate a Center of Mass (CoM) trajectory based on a predefined ZMP trajectory. Wieber \cite{wieber2006trajectory} improved the performance of this approach in the presence of relatively severe pushes, by recomputing the trajectories in a Model Predictive Control (MPC) framework. In these approaches a preview over several steps are considered for gait planning, but fixed step location and timing are assumed for this preview period.

 Besides optimization-based approaches for real-time walking pattern generation, analytical methods have been presented. In traditional analytical approaches, position and velocity of the CoM were employed to generate a trajectory for the CoM consistent with predefined footprints (ZMP trajectory) \cite{harada2006analytical}. However, by constraining the position and velocity of the CoM, both divergent and convergent parts of the LIPM dynamics are constrained. To remedy this, Takaneka et al. \cite{takenaka2009real} constrained the divergent part of the CoM to generate the Divergent Component of Motion (DCM) trajectory based on predefined footprints (ZMP trajectory). Before \cite{takenaka2009real}, the DCM had been used by Hof et al. \cite{hof2008extrapolated} to explain human walking properties under the name of extrapolated Center of Mass (XCoM), and by Pratt et al. \cite{pratt2006capture} with the name of Capture Point (CP), as the point on which the robot should step to come to rest.
 
By employing the DCM, Englsberger et al. \cite{englsberger2015three} proposed a method to control the unstable part of the CoM dynamics without affecting the stable part. Although this controller can react to the disturbances very fast, perfect DCM tracking needs unconstrained manipulation of the Center of Pressure (CoP). However, manipulating the CoP has different restrictions for robots with different structures. For instance, for robots with point contact feet (assuming instantaneous double support phase), it is not possible to modulate the CoP, while for robots with passive ankles the CoP cannot be manipulated directly by the ankle joints torques. In fact, changing the CoP location to control the DCM trajectory is limited to the robot structure, and in the best case can only be controlled directly inside the Support Polygon (SP), using a limited amount of ankle torque.

Contrary to the CoP manipulation for tracking the DCM, step adjustment is a more significant tool for stabilizing biped robots. The reason is that the next step location can be selected in a relatively large area compared to the SP. For example, for a point feet biped robot, the SP is the supporting contact point, while the next step location is only restricted by the kinematic range of the robot leg. Even for a robot with limited joints motion and on a very constrained environment, there exists an area for step adaptation which is larger than the contact point. As a result, step adaptation algorithms using analytical methods \cite{englsberger2015three} as well as optimization approaches \cite{diedam2008online,herdt2010online} have been suggested to make walking patterns more robust against disturbances. However, in all these methods the step timing is never adapted. The reason is that in order to keep the problem convex in a previewed number of steps, optimization based stepping pattern generators typically assume fixed step duration. Some works tried to use the step duration as an optimization variable\cite{aftab2012ankle,kryczka2015online}, however they resulted in a nonlinear optimization problem which is computationally expensive and also convergence to the global minimum is not guaranteed.

Compared to the CoP manipulation which is carried out at each control cycle, step adjustment is performed on a larger timing scale (each step time). In this paper, we show that by adjusting the step timing, we can make the step adjustment an even more effective tool for having robust gaits, without sacrificing computational efficiency. We propose a method for real-time adaptation of the step location and timing, based on DCM measurement. In our method, the problem is written as a quadratic program which can be solved in real-time. To demonstrate the robustness of our method, we conduct simulations in the presence of different disturbances. Comparison with state of the art preview controllers shows a significant robustness improvement when timing adaptation is added to step adjustment.

The rest of this paper is structured as follows: in section II, the proposed method for adjusting step location and timing is described. Section III deals with generating online swing foot trajectory. In section IV, simulation results are presented and discussed. Finally, section V concludes the findings.

\begin{figure}
\centering
\includegraphics[clip,trim=7cm 18.2cm 6cm 2.9cm,width=9cm]{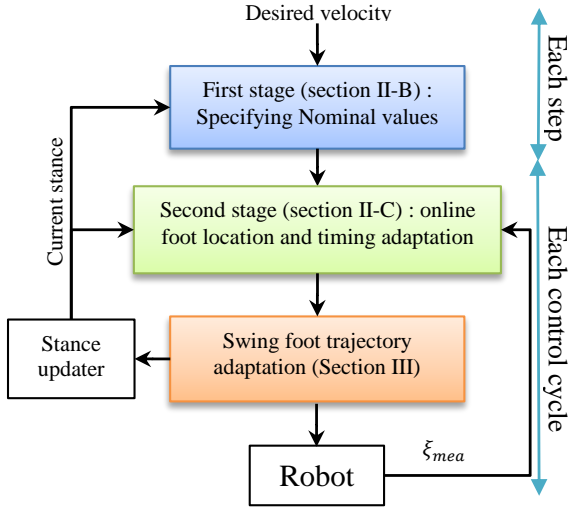}
\caption{Block diagram of the proposed method. In the first stage, at the start of each step the nominal values which are the step length and step duration are decided based on a desired velocity and consistent with the constraints. In the second stage which is carried out at each control cycle, the location and landing time of the swing foot are adapted through a small sized Quadratic Program (QP) optimization, using DCM measurement. To realize the adapted step length and step timing, the swing foot trajectory is regenerated at each control cycle.}
\vspace{-1.5em}
\label{block_diagram}
\end{figure}

\section{STEP LOCATION AND TIMING SELECTION}
As it is illustrated in Fig. \ref{block_diagram}, the two stages of our method for specifying the landing location and time of the swing foot are run at two different frequencies. The first stage is carried out at the start of each step and generates nominal values for the step location, step duration, and the desired offset between the landing point of the swing foot and the Divergent Component of Motion (DCM) at the end of the step (from now on we name this value "DCM offset"). Given the nominal values, the second stage generates the desired location and time for the swing foot landing at each control cycle, based on DCM measurement. Then, the outputs of this stage are exploited for generating swing foot trajectory (section III).

\subsection{Fundamentals}

The LIPM constrains motion of the Center of Mass (CoM) on a plane (horizontal plane for walking on a flat surface), by using a telescopic massless link connecting the CoP to the CoM \cite{kajita20013d}. The dynamics of this system may be formulated as:
\begin{equation}
\label{LIPM}
\ddot{x} = \omega_0^2 (x-u_0)
\end{equation}

in which $x$ is a 2-D vector containing CoM horizontal components (the vertical component has a fixed value $ z_0$), and $u_0$ is the CoP vector ($u_0=[CoP_x,CoP_y]^T$). Furthermore, $\omega_0$ is the natural frequency of the pendulum ($\omega_0=\sqrt[]{g/z_0}$, where $g$ is the gravity constant, and $z_0$ is the CoM height).

By considering CoM ($x$) and DCM ($\xi=x+\dot{x}/\omega_0$) as the state variables, the LIPM dynamics in the state space form may be specified as:
\begin{align}
\label{DCM_CoM}
\begin{split}
&\dot{x} = \omega_0 (\xi-x) \\ 
&\dot{\xi} = \omega_0 (\xi-u_0)
\end{split}
\end{align}

Equation (\ref{DCM_CoM}) decomposes the LIPM dynamics into its stable and unstable parts, where the CoM converges to the DCM and the DCM is pushed away by the CoP. Hence, in order to have a stable walking pattern, it is enough to constrain the DCM motion during walking, without restricting the other state of the system. Based on the second equation of (\ref{DCM_CoM}), one possibility for constraining the DCM motion is to constrain the DCM at the end of a limited time $\xi(T)$ \cite{englsberger2015three}. In fact in this case, the differential equation is solved as a final value problem and by having a limited value at the end of a specified time, the DCM trajectory converges. The other possibility is to instantaneously change the CoP location $u_0$ at a specified time (stepping). Stepping is used in this paper to constrain the DCM motion and stabilize the system, while the location and time of stepping are adapted using DCM measurement.

By solving the second equation of (\ref{DCM_CoM}) as an initial value problem, the DCM motion based on the natural dynamics of
the LIPM can be obtained:
\begin{equation}
\label{initial_value}
\xi(t) = (\xi_0-u_0) e^{\omega_0 t}+u_0
\end{equation}

The DCM at the end of a step is obtained by substituting the step duration $T$ in (\ref{initial_value}):
\begin{equation}
\label{initial@T}
\xi_T = (\xi_0-u_0) e^{\omega_0 T}+u_0
\end{equation}

Now, we define the DCM offset as:
\begin{equation}
\label{DCM_offset}
b=\xi_0-u_0=\xi_T-u_T
\end{equation}

where $u_0$ and $u_T$ are the current and next footprints,  respectively. Furthermore, $\xi_0$ and $\xi_T$ are the DCM at the start and end of the step. The DCM offset $b$ has a crucial role in our walking pattern generation method. By achieving this value at the end of a step, we make sure that the CoM travels a desired distance during a specified time in the next step (without perturbation) which realizes the desired average
velocity during $T$. As a result, by measuring the DCM we can control the next step location and step timing such that the desired offset is realized at the end of the step.

\subsection{First Stage: Specifying Nominal Values}

In the first stage of our method, we propose a procedure to find the gait variables consistent with a desired walking velocity, rather than predefining these variables \cite{englsberger2015three}. In fact, in this stage we aim at finding an optimum set of step length and step duration that satisfies the robot and environment constraints. Hence, the problem is to find the nominal step length ($L_{nom}$), step width ($W_{nom}$), and step duration ($T_{nom}$), subject to kinematic and dynamic constraints:
\begin{align}
\label{ineq&eq}
&v_x = \frac{L_{nom}}{T_{nom}} \nonumber \\
&v_y = \frac{W_{nom}}{T_{nom}}\nonumber \\
&L_{min} \leq L_{nom} \leq L_{max}\\
&W_{min} \leq W_{nom} \leq W_{max}\nonumber\\
&T_{min} \leq T_{nom} \leq T_{max}\nonumber
\end{align}
in which $v_x$ and $v_y$ are the desired average walking velocities in sagittal and lateral directions, respectively. The $max$ and $min$ indices show the bounds on the gait variables. It should be noted that the inequality constraints on step location are a combination of the robot (limited step length and width) and environment (limited area for stepping) limitations. Besides, there is a constraint on the minimum step timing which limits the acceleration of the swing foot, while the constraint on the maximum step timing is to limit very slow stepping.

Various performance criteria can be considered to find the best step length and step duration for a desired walking velocity. Though some studies have been carried out to address this for human walking \cite{kuo2005energetic,kuo2001simple}, it seems more practical to consider some other criteria for biped robots walking with different limitations compared to human. For instance, one possibility is to minimize fluctuations on the instantaneous CoM velocity in order to have a smooth walking pattern. In this paper, we chose nominal step length and width according to a robustness criteria, where we maximize distance to their bounds. Selecting the variables far from the boundaries yields more versatility to adapt them in real-time (second stage). Combining the equality and inequality constraints of (\ref{ineq&eq}) yields:
\begin{align}
\label{bounds}
B_l^1 = \frac{L_{min}}{|v_x|} \leq \; &T_{nom} \leq \frac{L_{max}}{|v_x|} = B_u^1\nonumber\\ 
B_l^2 = \frac{W_{min}}{|v_y|} \leq \; &T_{nom} \leq \frac{W_{max}}{|v_y|} = B_u^2\nonumber\\
B_l^3 =T_{min} \leq \; &T_{nom} \leq T_{max}= B_u^3
\end{align}

where $B_l^i$ and $B_u^i$ are the $i$'th lower bound and upper bound, respectively. It should be noted that the first two equations are valid for non-zero velocities. In the case where the speed in one of the directions is zero, the bounds on that direction are ignored, and the problem is solved with the rest of the equations. In order for all the inequality constraints of (\ref{bounds}) to be valid, the following equation should hold:
\begin{equation}
\label{final_bounds}
B_l=max(B_l^1,B_l^2,B_l^3) \leq T_{nom} \leq min(B_u^1,B_u^2,B_u^3)=B_u
\end{equation}

Hence, using equality equations of (\ref{ineq&eq}) and inequality equation of (\ref{final_bounds}), we can select the nominal step timing in the middle of the span and obtain the other variables as well:
\begin{align}
\label{nominal_values}
&T_{nom}=\frac{B_l+B_u}{2}\nonumber\\ 
&L_{nom}=v_x(\frac{B_l+B_u}{2})\\ 
&W_{nom}=v_y(\frac{B_l+B_u}{2})\nonumber
\end{align}

Consistent with these nominal values, we can compute the desired DCM offsets which realize these values in the absence of disturbances. Using (\ref {initial@T}) and (\ref {DCM_offset}), the nominal DCM offset in each direction for having nominal step length and width in nominal step duration can be obtained:
\begin{align}
\label{nominal_offset}
&b_{x,nom}=\frac{L_{nom}}{e^{\omega_0 T_{nom}}-1}\nonumber\\ 
&b_{y,nom}=(-1)^n\frac{l_p}{1+e^{\omega_0 T_{nom}}}- \frac{W_{nom}}{1-e^{\omega_0 T_{nom}}}
\end{align}
in which $l_p$ is the length of the robot pelvis (the default step width) and $n$ is the index for distinguishing the left and right feet.
\subsection{Second stage: online foot location and timing adaptation}
The second stage of our proposed algorithm which is carried out at each control cycle (Fig. \ref{block_diagram}) deals with adapting the gait characteristics based on DCM measurement. In this stage, we are given the nominal values of the step length, step time, and DCM offset. Our goal is to find the desired values for these variables as close as possible to the nominal values, based on the current measurement of the DCM. Hence, we define an optimization problem which automatically generates the desired values at each control cycle.

By solving the second equation of (\ref{DCM_CoM}) as a final value problem, we can write the LIPM solution in terms of the next footprint location, the step duration, and the DCM offset:
\begin{equation}
\label{final_value}
u_T = (\xi_{mea}-u_0) e^{\omega_0 (T-t)}+u_0-b  \quad ,  \quad  0 \leq t \leq T
\end{equation}

in which $\xi_{mea}$ is the measured DCM. Equation (\ref {final_value}) is a nonlinear equality with respect to the step time ($T$). As a result, the corresponding optimization problem would be nonlinear which introduces some issues for real-time applications \cite{herzog2015trajectory}. However, further inspection on (\ref{final_value}) reveals that by defining a new variable $\tau$ such that:
\begin{equation}
\label{variable_changing}
\tau=e^{\omega_0T} \Longrightarrow T=\frac{1}{\omega_0} log(\tau)
\end{equation}

We can relate (\ref {final_value}) in a linear equality form with respect to the variables:
\begin{equation}
\label{linear_equality}
u_T - (\xi_{mea}-u_0) e^{-\omega_0 t} \tau+b=u_0 \quad , \quad 0 \leq t \leq T 
\end{equation}

The concept behind this change of variable is that rather than dividing the solution space with respect to step duration into equal parts, we divide it in a logarithmic fashion. Then, after computing $\tau$ from the optimization problem, we can use the second equation of (\ref {variable_changing}) to obtain the step duration. Now that the main equality constraint is linear, we can write the problem as a Quadratic Program (QP) to yield the desired gait values as close as possible to the nominal values obtained from the first stage, and consistent with the constraints:
\begin{align}
\label{QP}
\underset{u_{T,x}, u_{T,y}, \tau, b_x, b_y}{\text{min.}} \alpha_1 \Vert u_{T,x}-L_{nom} \Vert + \alpha_2 \Vert u_{T,y}-W_{nom}\Vert  + \nonumber\\
 \alpha_3 \Vert \tau-\tau_{nom} \Vert + \alpha_4 \Vert b_x-b_{x,nom} \Vert  + \alpha_5 \Vert b_y-b_{y,nom} \Vert \nonumber\\ 
\text{s.t.} \quad \begin{bmatrix} 1 & 0 & 0 & 0 & 0 \\ -1 & 0 & 0 & 0 & 0\\ 0 & 1 & 0 & 0 & 0 \\ 0 & -1 & 0 & 0 & 0 \\0 & 0 & 1 & 0 & 0\\0 & 0 & -1 & 0 & 0	\end{bmatrix} 
\begin{bmatrix} u_{T,x} \\  u_{T,y} \\ \tau  \\ b_x  \\b_y 	\end{bmatrix}
\leq \begin{bmatrix} L_{max}\\-L_{min}\\W_{max}\\-W_{min}\\e^{\omega_0 T_{max}}\\-e^{\omega_0 T_{min}} \end{bmatrix}\nonumber\\
\begin{bmatrix} 1 & 0 & -(\xi_{mea}-u_0) e^{-\omega_0 t} & 1 & 0 \\ 0 & 1 & -(\xi_{mea}-u_0) e^{-\omega_0 t} & 0 & 1	\end{bmatrix} 
\begin{bmatrix} u_{T,x} \\  u_{T,y} \\ \tau  \\ b_x  \\b_y 	\end{bmatrix}= \begin{bmatrix} u_{0,x}\\u_{0,y} \end{bmatrix}            
\end{align}

The outputs of the optimization problem are the landing location and time of the swing foot as well as the DCM offset. It is important to note that the DCM offset should be an optimization variable because it needs to be adapted in case of large disturbances. This means, in some situations it is not possible to find a feasible set of step location and timing which realizes the desired DCM offset. However, since the DCM offset is related to the stability of the robot, we put a very high weight for it compared to the other terms. Doing so, the optimizer always tries to find a solution such that the DCM offset is realized (if possible). If the nominal DCM offset cannot be realized perfectly, then more than one step is required to converge back to the nominal walking speed, if the robot is capturable \cite{koolen2012capturability}.

\section{ONLINE FOOT TRAJECTORY GENERATION}

After specifying the step timing and location of the swing foot at the end of a step, the swing foot trajectory should be generated to realize the desired values. Since the next footprint and step timing can be changed at each control cycle, this procedure should be carried out in real-time. In fact, the trajectories for the swing foot should be regenerated at each control cycle to smoothly connect the desired swing foot state at the prior control cycle to the desired state at the end of step. We consider the polynomials such that the trajectories are continuous at the order of acceleration to ensure smooth control policies when used in an inverse dynamics control law \cite{herzog2016momentum}.

The boundary conditions for the trajectories of the swing foot in the horizontal plane ($x$ , $y$) can be written as:
\begin{align}
\label{horizontal_boundaries}
X(t_{k-1})=X_{k-1} \quad & \quad X(T)=u_T\nonumber\\ 
\dot X(t_{k-1})=\dot X_{k-1} \quad & \quad \dot X(T)=0\\ 
\ddot X(t_{k-1})=\ddot X_{k-1} \quad & \quad \ddot X(T)=0\nonumber
\end{align}

where index $k$ stands for the current sample time, and $X$ can be either sagittal or lateral component of the swing foot. In order to satisfy these constraints, we employ fifth order polynomials:
\begin{equation}
\label{horizontal_poly}
\sum_{i=0}^{5} c_it^i \quad , \quad t_{k-1} \leq t \leq T 
\end{equation}

in which $c_i$'s are the polynomials coefficients. Using the boundary conditions of (\ref{horizontal_boundaries}), the polynomials coefficients are obtained. Then, we just evaluate these polynomials for the current sample time ($t_k$) to obtain the desired value of the swing foot in the horizontal plane.

For the trajectory in the vertical direction, the problem is not straightforward. The reason is that the swing foot height increases to a certain value at a specified time (mid-time of the step), and then decreases to smoothly land on the ground. To generate a trajectory for the vertical component of the swing foot, we can use two fifth order polynomials for the two parts of the trajectory. For the first part, the trajectory connects the prior state to the state at the mid-time, while the second part connects the prior state to the state at the end time. However, this causes two main problems. First, change of step timing in the vicinity of the mid-time can cause a jump from the first part trajectory to the second part. Second, change of step timing in the second part generates unavoidable fluctuations in vertical direction which may cause the collision of the
swing foot with the ground.

Instead of two fifth order polynomials, we consider one 9th order polynomial for the whole step. The problem is to find the polynomials coefficients such that the swing foot height at the mid-time of the step is as close as possible to the desired height. Furthermore, the swing foot height during the step should be strictly positive and less than a maximum height. The polynomial coefficients are found by solving the following QP:
\begin{align}
\label{QP_poly}
\underset{c_i}{\text{min.}} &\quad \Vert Z(T/2)-Z_{des} \Vert \nonumber\\
\quad \text{s.t.} \quad & 0 \leq Z(t) \leq Z_{max}\nonumber\\
&Z(0)=0 \quad Z(t_{k-1})=Z_{k-1} \quad Z(T)=0\\ 
&\dot Z(0)=0 \quad \dot Z(t_{k-1})=\dot Z_{k-1} \quad \dot Z(T)=0\nonumber\\ 
&\ddot Z(0)=0 \quad \ddot Z(t_{k-1})=\ddot Z_{k-1} \quad \ddot Z(T)=0\nonumber            
\end{align}

In this equation $Z$ is the vertical component of the swing foot. By solving this equation, the polynomial coefficients $c_i$'s are obtained at each control cycle. Then, by evaluating the polynomial at current time, the desired foot trajectory is obtained in real-time.

\section{RESULTS AND DISCUSSIONS}
In this section, we present simulation results obtained from applying the proposed walking pattern generation method. In the first scenario, we present the results obtained from simulating the LIPM. We show that the model is able to recover from more severe pushes, when our controller with step timing adjustment is used. In the second scenario, we compare our controller with the state of the art of preview control \cite{herdt2010online} in terms of robustness against pushes. Finally, the third scenario demonstrates the effectiveness of our proposed method for controlling a full humanoid robot with passive ankles in a simulation environment.

\subsection{Simulation results using the LIPM}
In the first scenario, we consider the LIPM abstraction of a robot that is walking with a desired velocity. Footsteps and swing foot trajectories are computed as described in Section II and Section III. During each step, the stance foot is considered as the point of contact of the LIPM. At the end of a step where the point of contact of the LIPM is changed, the index of the foot $n$ is changed. By changing the foot index, the feasible area is computed for step location, based on the current state and index of the stance foot. We apply pushes on the robot and compare the recovery capabilities of our proposed controller to the case where no step timing adjustment is employed. The robot mass is 60 Kg and the fixed CoM height is considered 80 cm, while the pelvis length is 20 cm. The limits on step length, width, and time are specified in TABLE \ref{physical_properties}. The properties approximately mimic a human size humanoid robot.

\begin{table}[h]
\caption{PHYSICAL PROPERTIES OF THE MODEL}
\label{physical_properties}
\vspace{-1em}
\begin{center}
\begin{tabular}{|c|c|c|c|}
\hline
\it value & \it description & \it min & \it max\\
\hline
$L$ & Step length & -50 ($cm$) & 50 ($cm$)\\
\hline
$W$ & Step width & 10 ($cm$) & 40 ($cm$)\\
\hline
$T$ & Step duration & 0.2 ($sec$) & 0.8 ($sec$)\\
\hline
\end{tabular}
\end{center}
\end{table}

\begin{figure}
\centering
\subfloat[without time adjustment]{%
  \includegraphics[clip,trim=1.9cm 1.0cm 1cm 1cm,width=6cm]{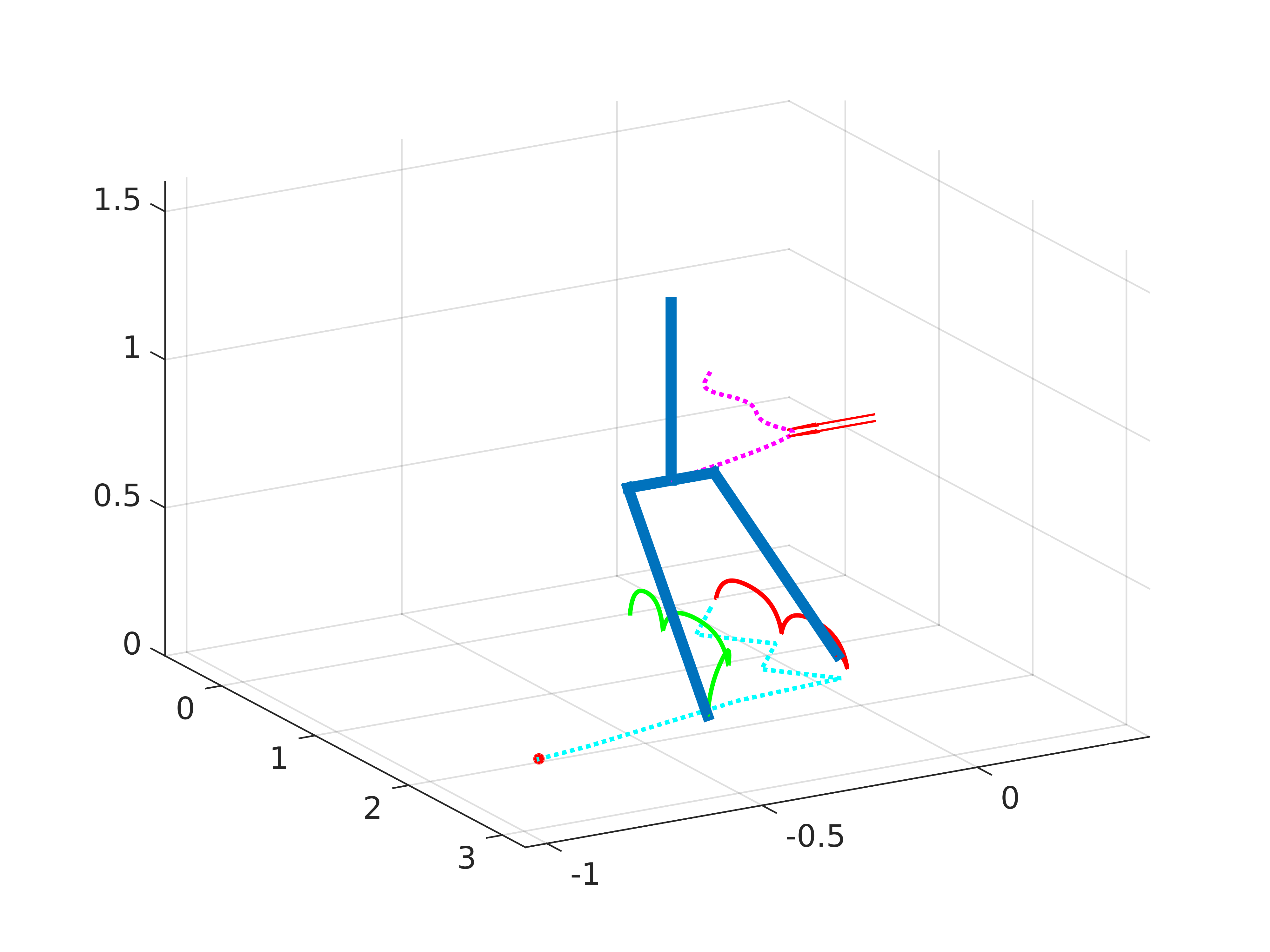}
}

\subfloat[with time adjustment]{%
  \includegraphics[clip,trim=1.4cm 1.cm .8cm 1.1cm,width=6cm]{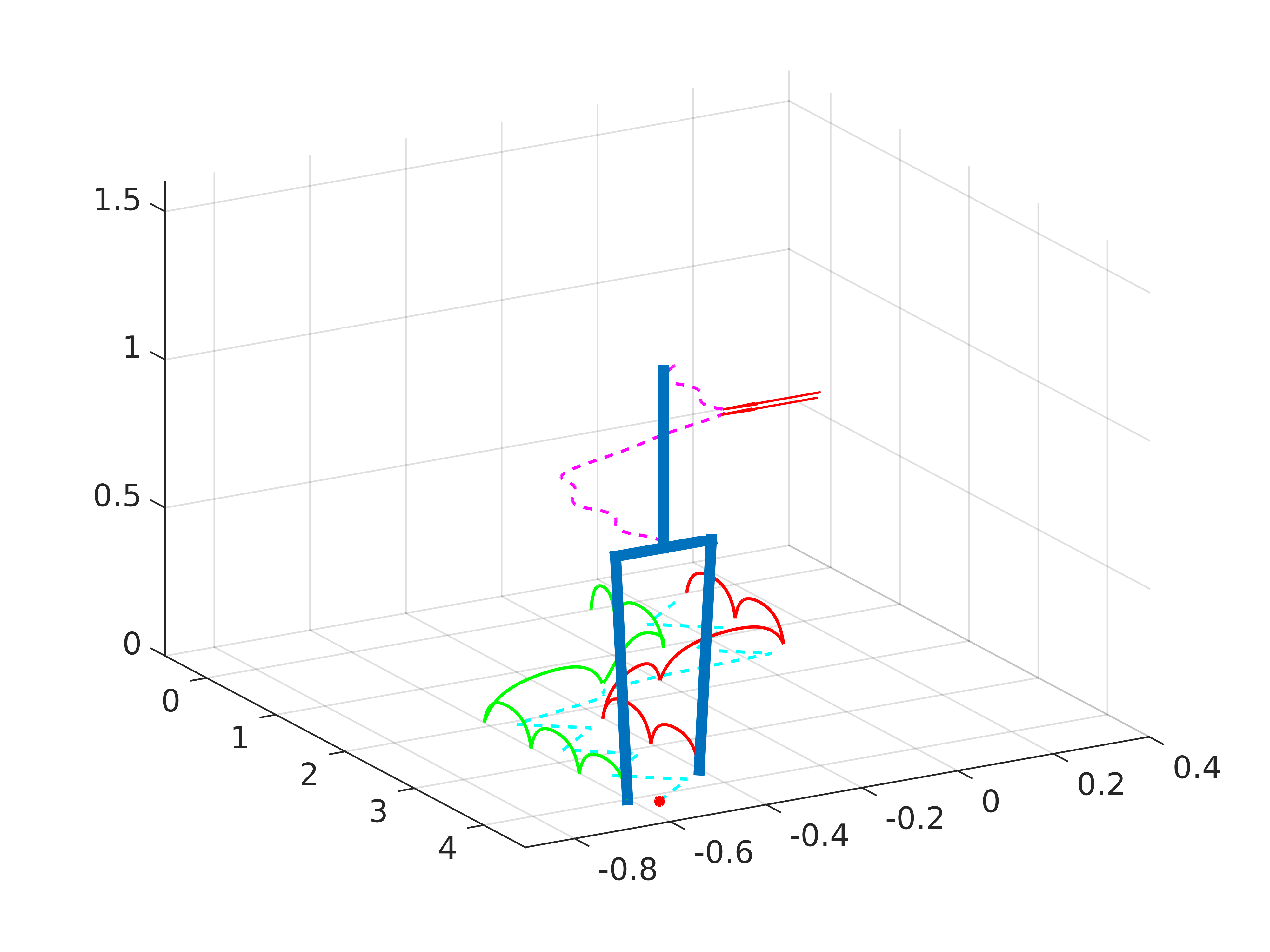}
}
\caption{Simulation (LIPM) of walking with the nominal speed of 1 m/s (green solid: right foot, red solid: left foot, magenta dashed: CoM, Cyan dashed: DCM projection on the ground). The robot is pushed at $t$=1.4 s to the right direction with a force $F$=325 N, during $\Delta t$ = 0.1 s. Step location is limited to a rectangular area and the time of stepping is limited with a minimum time (TABLE \ref{physical_properties}). case (a) (Left): The DCM diverges and the system cannot be captured without timing adjustment. Case (b) (Right): The system is able to recover from the push by making fast steps on the boundaries of feasible area, and continues its walking.}
\vspace{-1em}
\label{simulation}
\end{figure}

In this scenario, we compare our controller with one that uses fixed step durations $T_{nom}$. Resulting example gaits are visualized in Fig. \ref{simulation}. In this scenario, a velocity command ($v_x$=1 m/s) for forward walking is given. Based on the limitations specified in TABLE \ref{physical_properties}, the first stage of our proposed method generates the nominal step length and step duration using (\ref{nominal_values}), as well as the nominal DCM offset using (\ref{nominal_offset}). After four steps, the robot is pushed at $t$=1.4 s to the right direction with a force $F$=325 N, during $\Delta t$ = 0.1 s. We conduct two simulations to compare the results of fixed and optimized step durations. For the case (a) illustrated in Fig. \ref{simulation}, we employ (\ref{final_value}) for step  adjustment, using DCM measurement. In this case, the step duration is not adapted and has the nominal value during the motion. As it can be observed, the robot cannot recover from the push and the DCM diverges. It should be mentioned that if the step locations were not limited, the robot could recover from the push by arbitrarily changing the location of the next step.

In the case (b) in Fig. \ref{simulation}, we exploited the optimization procedure using (\ref{QP}) at each control cycle to generate the desired location and time of swing foot landing. In this case, for the first four steps where there is no disturbance, all the nominal values are realized. After the push in the fifth step, the robot employs a combination of step location and timing adjustment to recover from the push. In fact, the optimizer tries to find the values as close as possible to the nominal ones, while the DCM offset is given a very high weight. The result is such that after this relatively severe push, the robot makes three fast steps on the edges of the feasible area to recover from the disturbance. After the robot has recovered from the push, it continues its stepping with the desired velocity in forward direction. 

\begin{figure}
\centering
\setlength{\belowcaptionskip}{0mm} 
\includegraphics[clip,trim=5.5cm 5.8cm 10cm 1.cm,width=8.5cm,height=5cm]{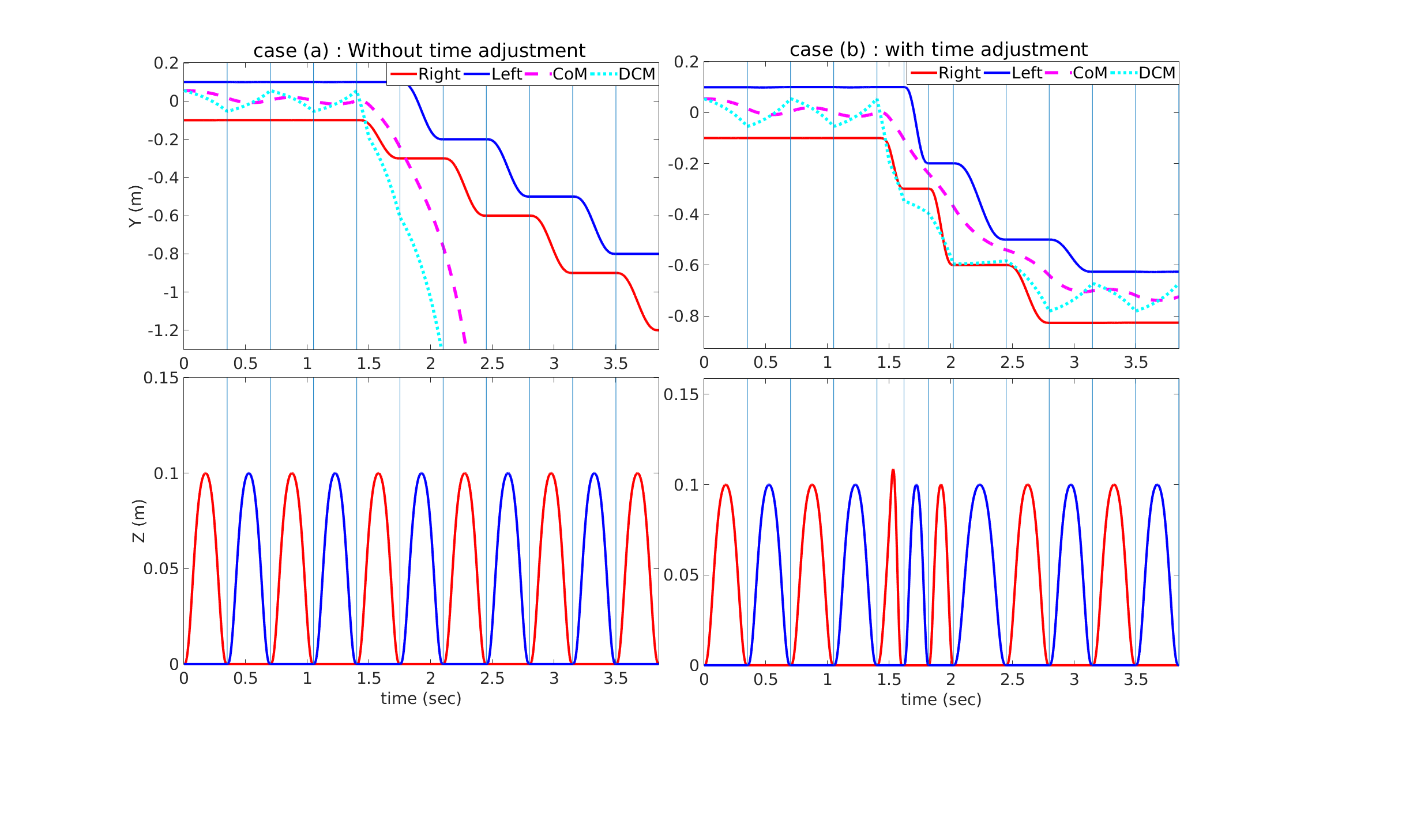}
\caption{Comparing the trajectories for the cases with and without time adjustment. The vertical lines show the step duration. All the components of the adapted swing foot trajectory are smooth in all directions. }
\vspace{-1.5em}
\label{CoMCoP}
\end{figure}

In Fig. \ref{CoMCoP}, the obtained trajectories for each case are demonstrated. The vertical lines show the step timing which is fixed for the case without step timing adjustment. As it can be observed in top figures, for the case without timing adjustment, the robot steps on the borders of feasible area to recover from the push. However, since in this case the stepping can be realized just in exact times, as time goes the DCM diverges and the swing foot is not able to capture the DCM fast. However, in the case with time adjustment, the algorithm adapts both next footprint and landing time based on the DCM measurement. As a result, in case of a severe push, the robot steps on the borders of the feasible area very fast to recover from the push. In the bottom figures, the swing foot trajectory in vertical direction for each case is shown. As it can be observed, employing the constrained problem of (\ref{QP_poly}), the swing foot trajectories in both cases are smooth and without undesired fluctuations.

\subsection{Comparison with \cite{herdt2010online}}
\begin{figure}
\centering
\setlength{\belowcaptionskip}{0mm} 
\includegraphics[clip,trim=.5cm .4cm 1.8cm 0.5cm,width=8.4cm,height=6.5cm]{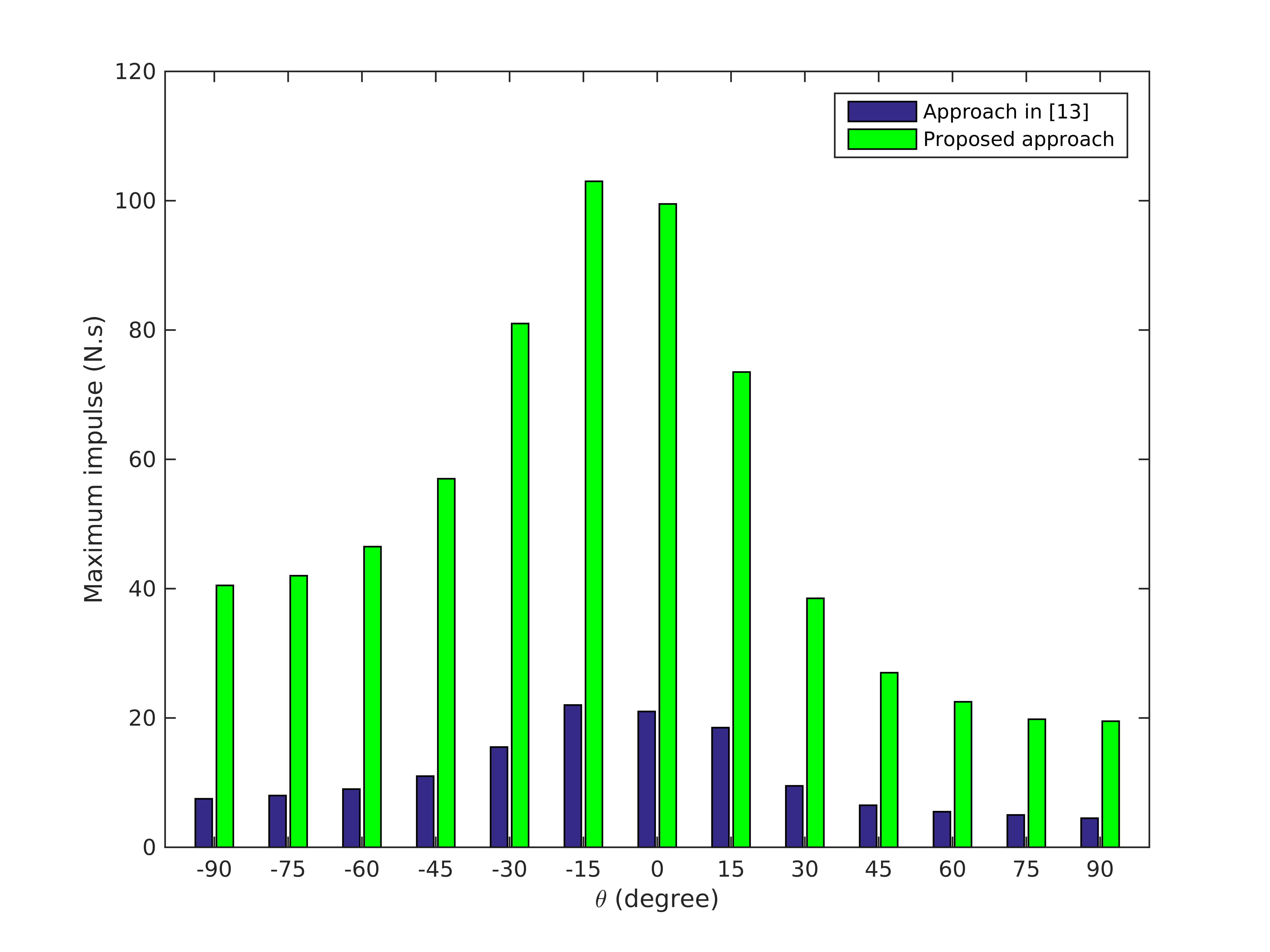}
\caption{comparison with the approach in [13]. $\theta$ is the angle between the direction of motion and the push direction (counterclockwise). Hence, $\theta$ =0 deg corresponds to a forward push, while $\theta$ =90 deg and $\theta$ =-90 deg represent pushes to the left and right directions, respectively. For each simulation, a force during $ \Delta t$ =0.1 s is applied at the start of a step in which the left foot is stance. As a result, since pushes to the left direction (positive $\theta$) impose the robot to self collision in the first step, step adjustment in this direction is more limited than the other direction. That is why the maximum push applied to the robot to the right direction (negative $\theta$ values) is more than the left direction (positive $\theta$ values).}
\vspace{-1.5em}
\label{comparison}
\end{figure}

In the second scenario, we compare the robustness of our proposed optimization procedure with time adjustment to the proposed approach in \cite{herdt2010online}. The proposed walking pattern generation approach in \cite{herdt2010online} and variations of this technique \cite{kajita2003biped,diedam2008online,wieber2006trajectory} are standard walking pattern generators in the literature. That is why we compare our results to this approach. The optimizer in \cite{herdt2010online} recomputes the CoM trajectory and footstep locations in real-time for a previewed period (typically 1.6 s) to realize a desired velocity of walking. We applied the same parameters for both approaches using an LIPM with point contact and computed the maximum push that each approach can recover from in various directions (Fig. \ref{comparison}). The value $\theta$ is the angle between the direction of motion and the push direction (counterclockwise). Hence, the negative values show the pushes to the right direction. We considered stepping with zero velocity to have symmetric results, while similar results can be obtained for different walking velocities. Therefore, the results are illustrated for -90 deg$\leq \theta \leq$ 90 deg, and for the other half symmetric results are obtained which are not shown. For each simulation, a force during $ \Delta t$ =0.1 s is applied at the start of a step in which the left foot is stance. It should be noted that for stepping in place with zero velocity $T_{nom}$=0.5 s is obtained from the first stage of our algorithm and considered for the step timing of both approaches.

As it can be observed in Fig. \ref{comparison}, our approach with time adjustment can recover from much more severe pushes compared to the approach in \cite{herdt2010online}. For the sideward pushes, the direction of the push affects the maximum value of the push that the robot is able to recover from. Because if the push is in the direction that with step adjustment the robot is prone to experience a self-collision (Left in this case with positive $\theta$ values), the feasible area is more limited than the other direction. Furthermore, the maximum push in the direction with $\theta =-15 \: deg$ (outer corner of the rectangle of feasible area) is more than the other directions, because the feasible area for stepping in this direction is larger than the other directions.  

Following points can be deduced from the LIPM simulation results and the comparison:

\begin{itemize}
\item \textit{Generality} In our proposed approach, we employed the step location and timing adjustment for stabilizing the stepping. In fact, without controlling the CoM or DCM at each control cycle, we could generate robust gaits. This suggests that our optimizer can be applied on biped robots with various structures, i. e. with active ankle or passive ankle or point contact foot. 

\item \textit{Robustness} For stepping with a nominal step duration, our experiments suggested that our approach is significantly more robust than an MPC approach with several preview steps but no timing adjustment. A problem of including timing duration in a receding horizon approach over several steps is that the optimization problem looses linearity (as in \cite{aftab2012ankle,kryczka2015online}). The reason is that the time evolution of the dynamics (\ref{initial_value}) multiplies the state at the beginning of a step ($\xi_0$ or $x_0$) with an exponential of time. Both multiplication of the optimization variables and the exponential term make the main equality constraint nonlinear. In our approach, we solve a convex optimization problem by only looking at the next step location and timing, and using the change of variable of (\ref{variable_changing}).

\item \textit{Nominal gait variables} We ran the optimizer in \cite{herdt2010online} with minimum step timing $T_{min}$, and obtained as same robustness as our approach. In fact, in the case of a very severe push, both algorithms yield stepping on the boundaries of the feasible area at the minimum step timing. The difference is that in our approach the stepping is done at the nominal timing as far as possible from boundaries. Then, in the case of a disturbance the step timing is adapted. However in this case, the optimizer in \cite{herdt2010online} works with the minimum step timing as a nominal behaviour. Stepping with minimum step timing as a nominal behaviour introduces many problems. For instance, the possibility of the system failure increases, or the energy efficiency decreases. 

\item \textit{Computational efficiency} Since in our optimization problem no horizon is considered, the size of the optimization problem drastically decreases compared to \cite{herdt2010online}. In the case of a severe push, both optimizers sacrifice the velocity tracking to recover from the push by stepping on the boundaries. In the case of an intermediate push, our optimizer sacrifices the velocity tracking to recover from the push as soon as possible. However, since the optimizer in \cite{herdt2010online} employs a horizon, it rejects the disturbance and also minimizes the velocity tracking error.

\end{itemize}

\subsection{Full humanoid simulation}
\begin{figure}
\centering
\setlength{\belowcaptionskip}{0mm} 
\subfloat[real robot]{%
  \includegraphics[width=2.5cm,height=5cm]{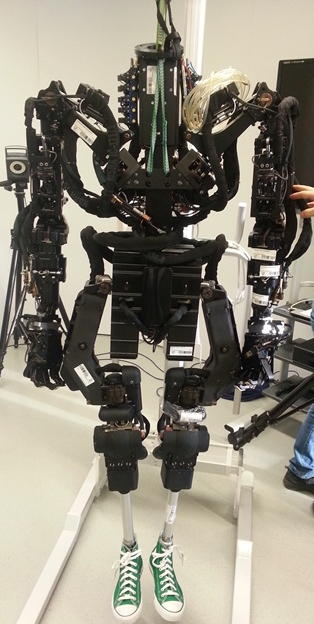}
}
\subfloat[simulated model]{%
  \includegraphics[width=2.5cm,height=5cm]{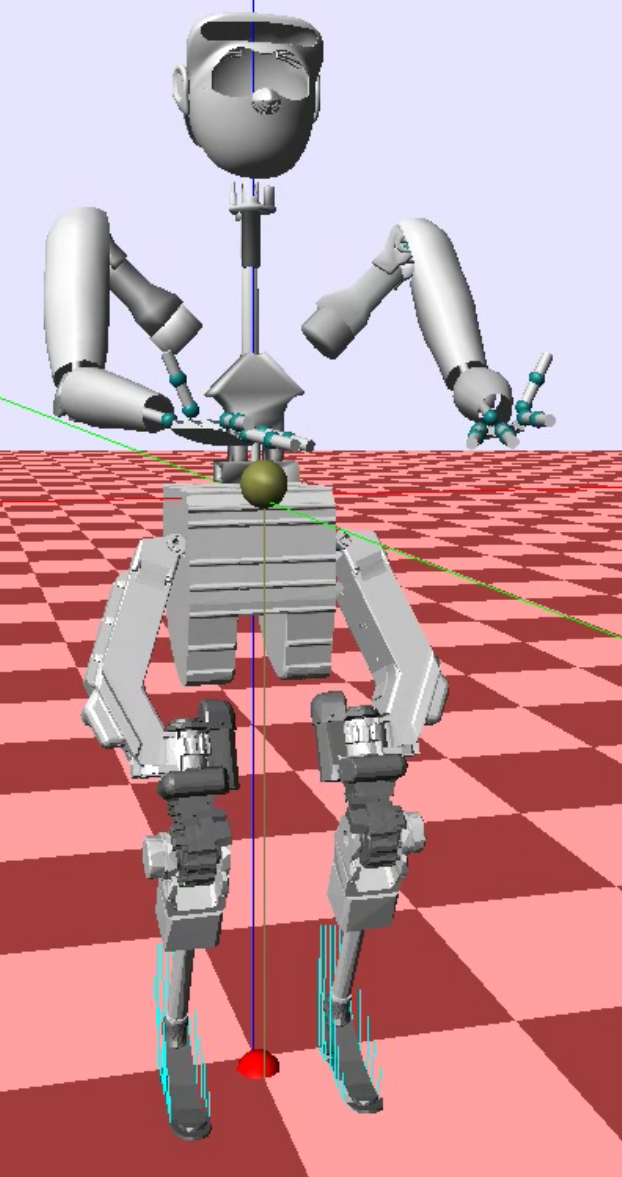}
}
\caption{The Sarcos humanoid robot Athena with passive ankles and prosthetic feet}
\vspace{-1.5em}
\label{athena}
\end{figure}

In the third scenario, we use our walking controller on a simulation of the Sarcos humanoid robot Athena in the SL simulation environment (Fig. \ref{athena})  \footnote{A summary of full humanoid simulation on different scenarios is on \url{https://www.youtube.com/watch?v=RVbCFm8V_DY}}. Each leg of the robot has 4 active degrees of freedom with passive ankle joints and prosthetic feet. For simulating the passive ankle joints, stiff springs and dampers are used. For the low-level control, we use a hierarchical inverse dynamics \cite{herzog2016momentum} to generate desired joint torques for a set of desired tasks \cite{khadiv2016stepping}. 

Since the robot ankle joints are passive, explicitly controlling the CoM or the DCM is problematic. The reason is that unconstrained manipulation of the CoP inside the support polygon for controlling the CoM or the DCM needs large arms and upper-body motions. This is not a desired behavior during a normal walking. As a result, our controller which uses DCM measurement to adjust swing foot landing location and time is a useful tool for stabilizing the robot walking. In the hierarchical inverse dynamics, we put the stance foot control and the CoM height control tasks in the first rank of the task hierarchy. The swing foot control, the posture control and the force regularization tasks are put respectively in the second, third and fourth rank. 

\begin{figure}
\centering
\setlength{\belowcaptionskip}{0mm} 
\captionsetup[subfigure]{labelformat=empty,position=top}
\subfloat[t=1 s]{%
  \includegraphics[width=1.5cm,height=2.1cm]{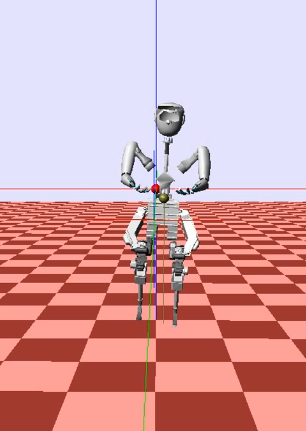}
}
\subfloat[t=2 s]{%
  \includegraphics[width=1.5cm,height=2.1cm]{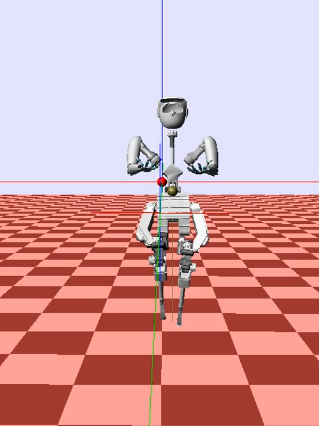}
}
\subfloat[t=3 s]{%
  \includegraphics[width=1.5cm,height=2.1cm]{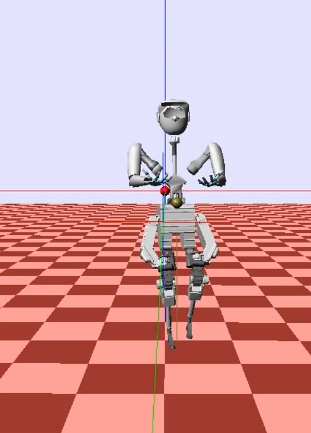}
}
\subfloat[t=4 s]{%
  \includegraphics[width=1.5cm,height=2.1cm]{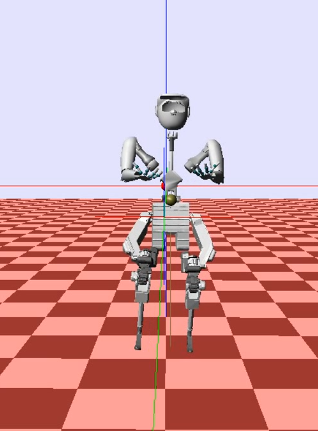}
}

\subfloat[t=5 s]{%
  \includegraphics[width=1.5cm,height=2.1cm]{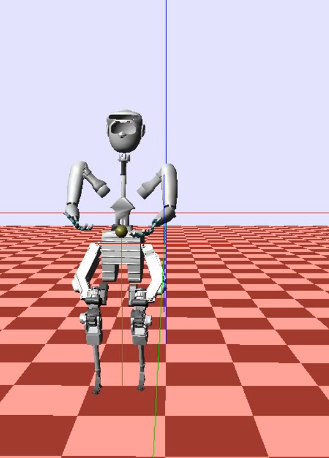}
}
\subfloat[t=6 s]{%
  \includegraphics[width=1.5cm,height=2.1cm]{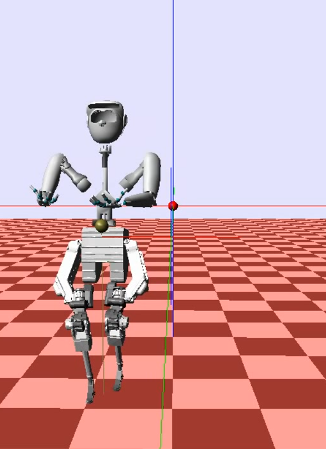}
}
\subfloat[t=7 s]{%
  \includegraphics[width=1.5cm,height=2.1cm]{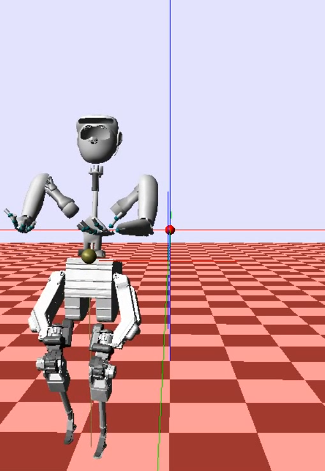}
}
\subfloat[t=8 s]{%
  \includegraphics[width=1.5cm,height=2.1cm]{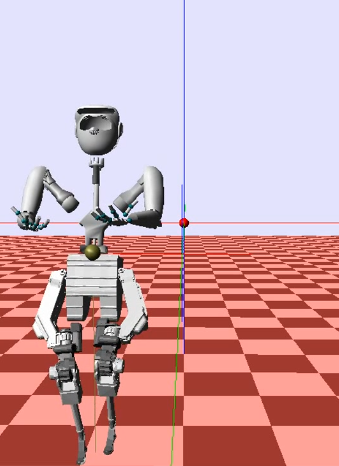}
}
\caption{Athena walks forward with $v_x$=0.2 m/s in the SL simulation environment. The robot is pushed at sixth step $t$=3.7 s by $F$=200 N during $\Delta t$= 0.1 s.}
\vspace{-1.5em}
\label{athena_simulation}
\end{figure}

Figure \ref{athena_simulation} visualizes a forward walking simulation ($v_x$=0.2 m/s). In this simulation, based on the commanded velocity, the first stage of our method computes the nominal gait parameters. The constraints on the step location and timing are the same as the first scenario (specified in TABLE \ref{physical_properties}). Then, the second stage adapts the gait parameters using DCM measurement. Since the robot has finite size feet, we use the measured CoP from the interacting forces as the current contact point in our controller. Finally, the swing foot trajectory is regenerated at each control cycle to adapt the landing point. In this simulation scenario, the robot is pushed to the right by a force  $F$=200 N, at $t$=3.7 s during $\Delta t$= 0.1 s. To recover from the push, the robot starts stepping to the right direction. Then, once the push is rejected, the robot continues its nominal forward walking. 

\begin{figure}
\centering
\includegraphics[clip,trim=1cm .4cm 1.6cm 0.5cm,width=8cm]{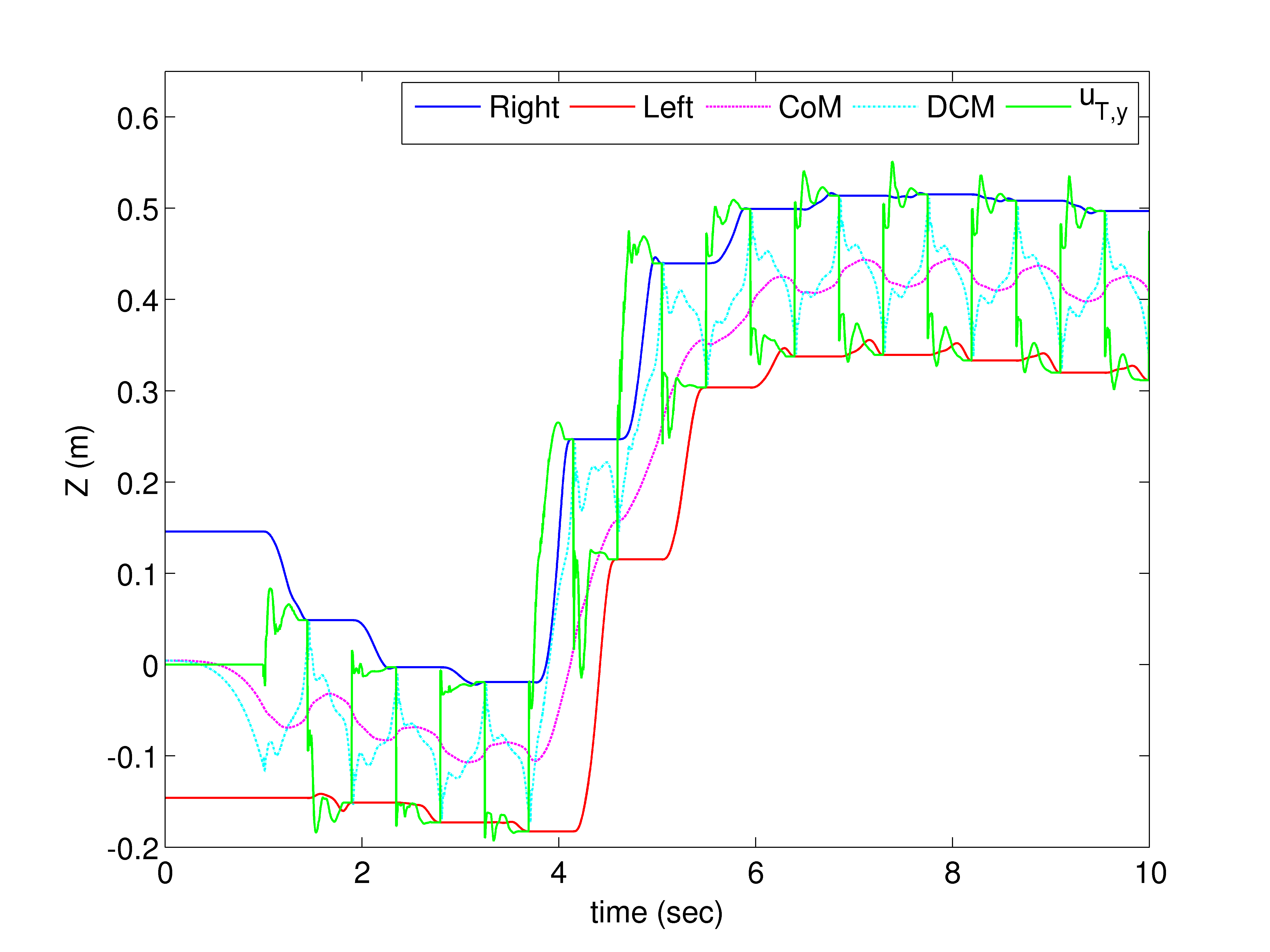}
\caption{The lateral trajectories during the third simulation scenario. The desired lateral velocity is zero during this forward walking simulation scenario. However, when the push is exerted, the controller sacrifices lateral velocity tracking to recover the robot from the push. }
\label{CoMCoP_sim}
\end{figure}

\begin{figure}
\centering
\includegraphics[clip,trim=1.cm .4cm 1.6cm 1cm,width=8.5cm]{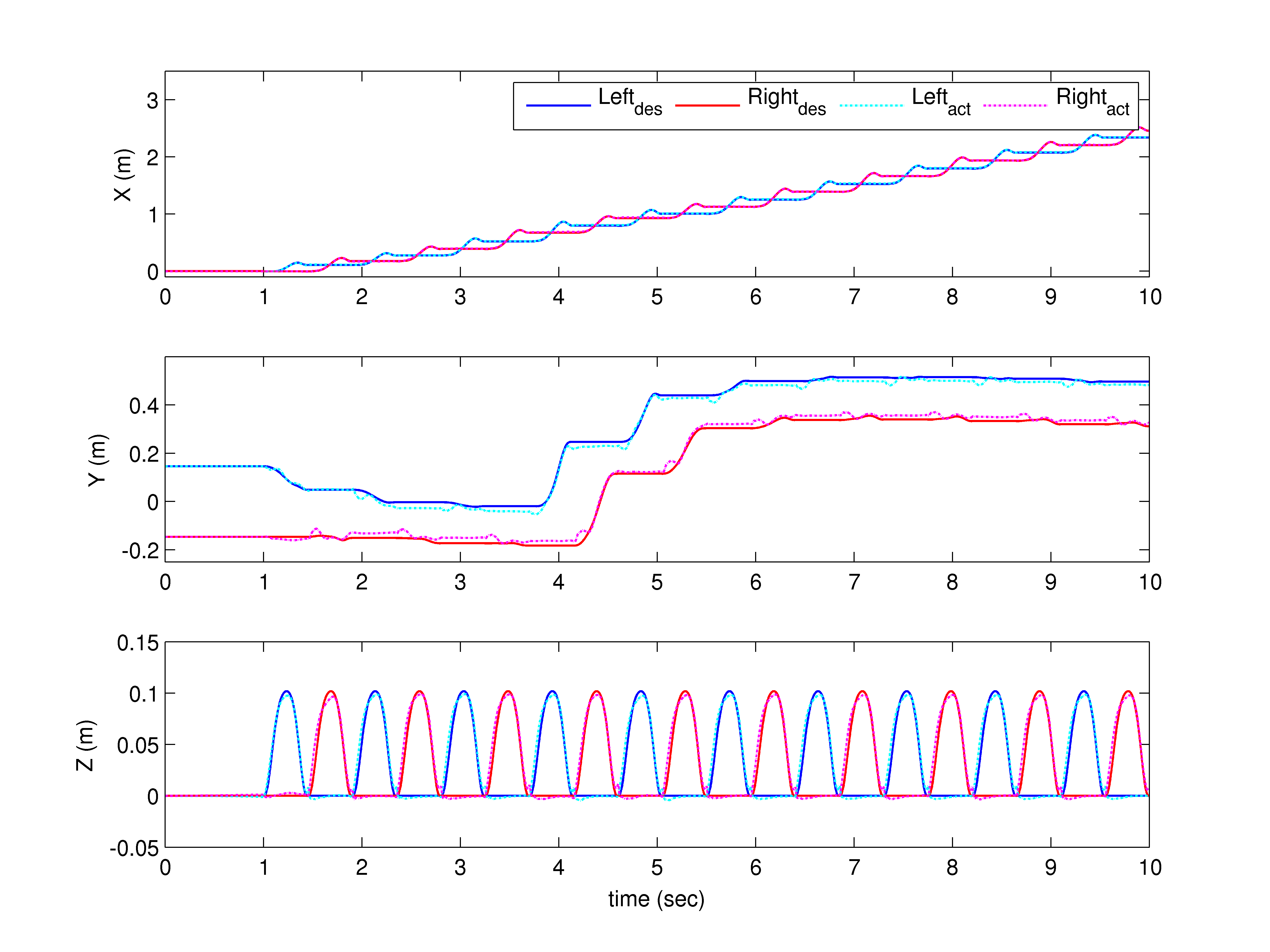}
\caption{The desired and actual feet trajectories. The low-level controller tracks the smooth feet trajectories generated by our real-time walking controller. }
\vspace{-0.5em}
\label{foot}
\end{figure}

Figure \ref{CoMCoP_sim} illustrates the trajectories in the lateral direction for this simulation scenario. In this figure, the updated swing foot landing point $u_{T,y}$ at each control cycle is shown. As it can be seen, the feet trajectories are adapted to realize the desired landing locations. When the push is applied, the controller sacrifices the lateral velocity tracking and adjusts the gait parameters to recover the robot. Figure \ref{foot} shows the desired and actual feet trajectories. As it can be observed in this figure, the adapted trajectories of the feet are smooth, and the controller is able to track the trajectories.

\section{CONCLUSIONS}
In this paper, we proposed a method for generating robust walking patterns for biped robots. The first stage of this method computes the best step location and duration, for a specified walking velocity at the start of each step. Then, the second stage adapts these values using DCM measurement, such that the desired DCM offset which is related to the stability is realized, and the gait properties are as close as possible to the nominal values. To adapt the updated gait variables, the swing foot trajectory is regenerated at each control cycle in real-time. Different simulations demonstrated that the proposed approach is significantly more robust compared to the case where step adjustment without timing adjustment is employed, even when the preview controller is allowed several preview steps.

\bibliography{Master}
\bibliographystyle{IEEEtran}

\end{document}